\pdfoutput=1

\documentclass[11pt]{article}

\usepackage[]{EMNLP2023}

\usepackage{times}
\usepackage{latexsym}
\usepackage{graphicx}
\usepackage{booktabs}
\usepackage{tabularx}
\usepackage{url}
\usepackage[T1]{fontenc}
\usepackage[utf8]{inputenc}
\usepackage{microtype}
\usepackage{inconsolata}
\usepackage[normalem]{ulem}
\usepackage{subcaption}

\usepackage{xcolor}
\usepackage{fancyvrb}

\DefineVerbatimEnvironment{Highlighting}{Verbatim}{commandchars=\\\{\}}
\newenvironment{Shaded}{}{}

\newcommand{\ImportTok}[1]{\textcolor[rgb]{0.00,0.50,0.00}{\textbf{#1}}}

\newcommand{\KeywordTok}[1]{\textcolor[rgb]{0.00,0.44,0.13}{\textbf{#1}}}
\newcommand{\NormalTok}[1]{#1}
\newcommand{\OperatorTok}[1]{\textcolor[rgb]{0.40,0.40,0.40}{#1}}

\newcommand{\StringTok}[1]{\textcolor[rgb]{0.25,0.44,0.63}{#1}}

\usepackage{tikz}
\usetikzlibrary{decorations.pathreplacing}

\title{\texttt{pose-format}: Library for Viewing, Augmenting,
and Handling \texttt{.pose} Files}

\author{
Amit Moryossef\\
University of Zürich\\
\texttt{amitmoryossef@gmail.com}
\\\And
Mathias Müller\\
University of Zürich\\
\texttt{mmueller@cl.uzh.ch}\\
\newline\\
\url{https://github.com/sign-language-processing/pose}\\
\\\And
Rebecka Fahrni\\
University of Zürich\\
\texttt{rebecka.fahrni@uzh.ch}
}

\begin{document}
\maketitle

\begin{abstract}
Managing and analyzing pose data is a complex task, with challenges ranging from handling diverse file structures and data types to facilitating effective data manipulations such as normalization and augmentation. This paper presents \texttt{pose-format}, a comprehensive toolkit designed to address these challenges by providing a unified, flexible, and easy-to-use interface. The library includes a specialized file format that encapsulates various types of pose data, accommodating multiple individuals and an indefinite number of time frames, thus proving its utility for both image and video data. Furthermore, it offers seamless integration with popular numerical libraries such as NumPy, PyTorch, and TensorFlow, thereby enabling robust machine-learning applications. Through benchmarking, we demonstrate that our \texttt{.pose} file format offers vastly superior performance against prevalent formats like OpenPose, with added advantages like self-contained pose specification. Additionally, the library includes features for data normalization, augmentation, and easy-to-use visualization capabilities, both in Python and Browser environments. \texttt{pose-format} emerges as a one-stop solution, streamlining the complexities of pose data management and analysis.
\end{abstract}

\section{Introduction}
Working with pose data introduces many complexities, from the diversity in file structures to the variety of data types that need to be accommodated. Developers and researchers often find themselves juggling numerous data manipulation tasks such as normalization, augmentation, and visualization. In addition to these challenges, pose data itself can be inherently multidimensional, frequently encompassing multiple individuals and varying time frames. This creates an intricate ecosystem of variables that can be challenging to manage and analyze effectively, which is particularly important in fields like Sign Language Processing.

To overcome these complexities, we designed \texttt{pose-format}, a comprehensive toolkit to alleviate these challenges by offering a unified, flexible, and easy-to-use interface for managing and analyzing pose data. Designed with versatility in mind, the library includes a specialized file format that accommodates an array of pose types, multiple people, and an indefinite number of time frames, making it highly adaptable for both video and single-frame data. Users can effortlessly import \texttt{.pose} files and perform a range of manipulations such as data normalization and augmentation. The library also integrates seamlessly with popular numerical libraries like NumPy \citep{Harris2020ArrayPW}, PyTorch \citep{pytorch}, and TensorFlow \citep{tensorflow2015-whitepaper}, allowing for additional computational flexibility for machine learning. With features for easy visualization and compatibility with other popular pose data formats like OpenPose \citep{pose:cao2018openpose} and MediaPipe Holistic \citep{MediaPipe2020holistic}, the \texttt{pose-format} library emerges as a one-stop solution for all pose data management needs.

\begin{figure}[t]
    \centering
    \begin{tikzpicture}
    \node[anchor=west,opacity=0.5] at (0,0) {\includegraphics[width=2cm,height=2cm]{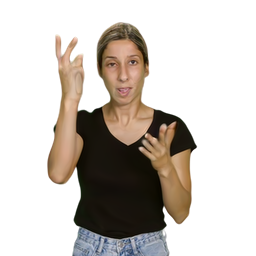}};
    \node[anchor=west,opacity=0.7] at (-0.5,-0.5) {\includegraphics[width=2cm,height=2cm]{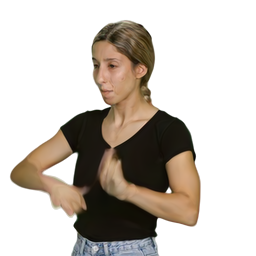}};
    \node[anchor=west] at (-1,-1) {\includegraphics[width=2cm,height=2cm]{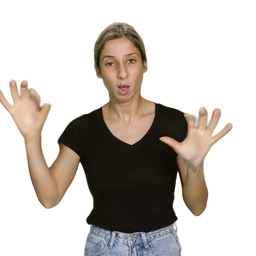}};
     \draw[->] (2.2,-0.5) -- node[below] {} (2.8,-0.5);
    
    \node[anchor=west,opacity=0.5] at (4,0) {\includegraphics[width=2cm,height=2cm]{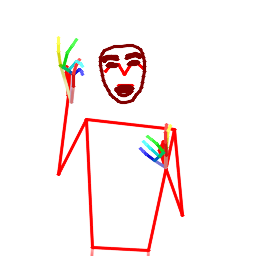}};
    \node[anchor=west,opacity=0.7] at (3.5,-0.5) {\includegraphics[width=2cm,height=2cm]{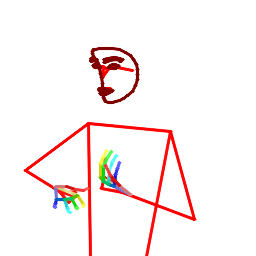}};
    \node[anchor=west,opacity=1] at (3,-1) {\includegraphics[width=2cm,height=2cm]{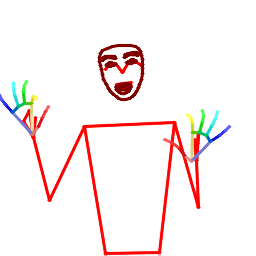}};
    
    \end{tikzpicture}
    \caption{Examples of human skeletal poses extracted from a sign language video sequence.}
    \label{fig:pose_example_vis}
\end{figure}

\section{Background}
\label{sec:background}

In the context of our library, a \emph{pose} consists of \emph{keypoints}, which are 2D or 3D coordinates marking points of interest usually on a human body in image or video frames (Figure \ref{fig:pose_example_vis}). Systems like OpenPose and MediaPipe Holistic are prominent for pose estimation but have differing methodologies and keypoint configurations. OpenPose, for instance, uses a classification objective and outputs 135 or 137 keypoints with 2D coordinates. MediaPipe Holistic employs a regression objective, estimating 543 keypoints with 3D coordinates.

Keypoints are hierarchically organized, often attached to larger body components like \texttt{LEFT HAND} or \texttt{FACE}. Moreover, models implicitly define which keypoints are connected, forming an underlying graph structure. 
Confidence metrics vary across systems. OpenPose assigns a confidence score to each classification, while MediaPipe Holistic only predicts the likelihood of each \texttt{BODY} keypoint's presence in the original image.

The utility of human pose estimation \citep{zheng2023deep} spans various fields such as human-computer interaction, motion capture, motion analysis, and mixed reality, with specialized applications like automatic sign language processing \citep{moryossef2021evaluating,muller-etal-2022-findings}.

\section{Justification}

\texttt{pose-format} addresses a void in the ecosystem by delivering a uniform layer of abstraction over disparate pose estimation system outputs, such as OpenPose and MediaPipe Holistic. The necessity for this unified interface arises from three primary factors: inconsistent standards, inadequacy of existing libraries, and performance bottlenecks.

\paragraph{Inconsistent Standards}

As delineated in \S\ref{sec:background}, there are competing pose estimation systems, each adhering to its own data storage and representation scheme. This inconsistency impedes interoperability between systems and makes the data hard to share or transition across different platforms. \texttt{pose-format} remedies this by standardizing how pose data is managed, making it simpler to operate with multiple systems, switch between them, or even disseminate pose data.

\paragraph{Limitations of Existing Libraries}

Current libraries focus extensively on low-level operations, lacking the higher-level abstractions that can expedite routine tasks. For instance, in the absence of our toolkit, users have to micromanage array values, discerning between coordinates and confidence scores or handling missing keypoints. Such intricacies detract from productivity and introduce unnecessary complexity. Our library fills this gap by offering user-friendly operations, many of which are indispensable for machine learning research, such as frame rate interpolation, rotation, scaling, frame dropout, or converting the underlying data into tensors of a specific machine learning library.

\paragraph{Efficiency}

As demonstrated in \S\ref{sec:benchmarking}, prevailing methods for pose data management suffer from performance limitations in both speed and storage. These inefficiencies create bottlenecks for data-intensive tasks, especially those prevalent in machine learning pipelines. \texttt{pose-format} offers optimized data storage and retrieval, mitigating these inefficiencies.

\section{Format Specification}

The core of the \texttt{pose-format} library is its specialized file format that accommodates a wide range of scenarios. This unique format enables the storing of multidimensional data capturing various pose types, multiple individuals, and an indefinite number of time frames. Currently, at version $0.1$, the file format is bifurcated into two components: the \texttt{Header} and the \texttt{Body}.

\subsection{Header (\texttt{PoseHeader})}

The header contains meta information that defines the overall structure of the pose data. This information is useful for visualization and code readability.
Specifically, it includes:

\paragraph{(\texttt{float32}) Version} The version of the file format.
\paragraph{(\texttt{uint16[3]}) Dimensions} Width, height, and depth specifications.
\paragraph{(\texttt{uint16}) Number of Components} The number of pose components.
\paragraph{Component Details} Each component includes its (\texttt{string}) name, (\texttt{string}) format, and the (\texttt{uint16}) number of points, (\texttt{uint16}) limbs, and (\texttt{uint16}) colors it contains.
\begin{itemize}
    \item (\texttt{string[]}) Names of points.
    \item (\texttt{uint16[2][]}) Start and end indices of limbs.
    \item (\texttt{uint16[3][]}) Points color RGB values.
\end{itemize}

\subsection{Body (\texttt{PoseBody})}

The body of the file comprises the actual pose data and includes the following:

\paragraph{(\texttt{uint16}) FPS} The frame rate of the pose.
\paragraph{(\texttt{uint16}) Number of frames} \sout{deprecated due to challenges for longer pose sequences.}
\paragraph{(\texttt{uint16}) Number of People} The number of people included in every frame.
\paragraph{(\texttt{float[][][][]}) Data} The coordinate of every point for every person in every frame.
\paragraph{(\texttt{float[][][]}) Confidence} The confidence for every point for every person in every frame.

This format's granularity and modularity make it aptly suited for a wide range of applications, from simple image-based pose representation to more complex video analysis tasks. By leveraging this detailed yet flexible format, the \texttt{pose-format} library ensures ease of use without sacrificing the intricacies that pose data often necessitates.

\subsection{v0.1 Limitations}

While the \texttt{pose-format} library has been designed to cater to a wide array of needs, there are some limitations and criticisms in the current file format that users should be aware of:
\begin{itemize}
    \item \textbf{FPS Representation}: The FPS is stored as \texttt{uint16}, which does not allow for floating-point values.
    \item \textbf{Number of Frames}: The number of frames is also restricted to \texttt{uint16}, which limits the frame count to 65,535. The current workaround calculates the number of frames based on the file size, which introduces computational overhead.
    \item \textbf{Pose Data Precision}: The pose data utilizes 32-bit floating-point values for storage. However, 16-bit floating-point numbers could be sufficient for many applications. Support for both types would improve memory efficiency.
    \item \textbf{Confidence Precision}: Similar to the pose data, the confidence metrics are stored as 32-bit floating-point numbers. A 16-bit representation would be more than sufficient for most practical purposes.
\end{itemize}

\section{Data Manipulations}

One of the key advantages of this toolkit is its robust support for various data manipulation tasks, which are crucial for the preprocessing and augmentation of pose data. This section elaborates on how the library facilitates operations such as normalization and augmentation.

\paragraph{Normalization}
Normalization is a crucial step to make pose data scale and translation invariant, thereby improving the effectiveness of downstream tasks like training machine learning models. Our toolkit offers a simple yet powerful interface to normalize pose data. For example, when dealing with human body poses, we can specify the names of the left and right shoulders, and the skeleton will be scaled such that the mean distance between the shoulders is equal to $1$, and the center point lies on $(0,0)$. If we deal with 3D poses, we can also specify a plane by naming three points, to make sure they always fall on the same plane. These normalizations remove the effect of camera angles and distance from the subject.

\paragraph{Augmentation}
Data augmentation is a technique to artificially increase the size and diversity of your training dataset by applying various transformations. In the context of pose data, these can include affine transformations such as translation, scaling, reflection, rotation, and shear, interpolation of frames at variable speeds, noise, and dropout, to name a few. 
The \texttt{pose-format} toolkit provides built-in functions to perform these augmentations effortlessly. You can either apply these transformations individually or chain them together to create a complex augmentation pipeline, thereby enhancing the library's adaptability to various project needs.

\paragraph{Integration with Numerical Libraries}

Data manipulations are seamlessly integrated with popular numerical libraries like NumPy, PyTorch, and TensorFlow. This facilitates easy data flow between data manipulation and machine learning models, reducing the friction in the data science pipeline. It allows loading and augmenting the data in a framework of your choosing, minimizing the number of memory copy operations.

\begin{table*}[t]
\centering
\begin{tabular}{@{}lccccc@{}}
\toprule
\# Frames & \multicolumn{2}{c}{\texttt{OpenPose}} & \multicolumn{3}{c}{\texttt{pose-format}} \\
\cmidrule(lr){2-3} \cmidrule(lr){4-6} 
 & Size & Speed & Size & Speed & Speed (Body) \\
\midrule
1       & $3.9\, \texttt{KB}$ & $37.4\, \mu \texttt{s} \pm 600\, \texttt{ns}$ & $3.6\, \texttt{KB}$ & $535\, \mu \texttt{s} \pm 66.1\, \mu \texttt{s}$ & $61.7\, \mu \texttt{s} \pm 6.94\, \mu \texttt{s}$ \\
10      & $38\, \texttt{KB}$ & $364\, \mu \texttt{s} \pm 6.9\, \mu \texttt{s}$ & $18\, \texttt{KB}$ & $490\, \mu \texttt{s} \pm 63.8\, \mu \texttt{s}$ & $57.9\, \mu \texttt{s} \pm 2.56\, \mu \texttt{s}$ \\
100     & $388\, \texttt{KB}$ & $3.75\, \texttt{ms} \pm 113\, \mu \texttt{s}$ & $163\, \texttt{KB}$ & $415\, \mu \texttt{s} \pm 49.7\, \mu \texttt{s}$ & $72.4\, \mu \texttt{s} \pm 4.87\, \mu \texttt{s}$ \\
1,000   & $3.9\, \texttt{MB}$ & $43.1\, \texttt{ms} \pm 704\, \mu \texttt{s}$ & $1.6\, \texttt{MB}$ & $658\, \mu \texttt{s} \pm 110\, \mu \texttt{s}$ & $228\, \mu \texttt{s} \pm 9.09\, \mu \texttt{s}$ \\
10,000  & $39\, \texttt{MB}$ & $439\, \texttt{ms} \pm 29.5\, \texttt{ms}$ & $16\, \texttt{MB}$ & $2.72\, \texttt{ms} \pm 110\, \mu \texttt{s}$ & $2.71\, \texttt{ms} \pm 245\, \mu \texttt{s}$ \\
\bottomrule
\end{tabular}
\caption{Benchmarking \texttt{pose-format} against \texttt{OpenPose} from the Public DGS Corpus. We compare both the resulting file size, and file read speed. \emph{Speed (Body)} measures loading the \emph{.pose} files data only, without metadata.}
\label{table:benchmarking}
\end{table*}

\section{Visualization}

The ability to visualize pose data is crucial for understanding its characteristics, debugging algorithms, and even for presentation purposes.

\paragraph{Python} In Python, users can make use of the \texttt{PoseVisualizer} class for different visualization tasks, such as visualizing the pose by itself as a sequence of still images, a video, a GIF, with the background being either a fixed color or overlayed on another video. An example of visualizing the pose as a video would be:

\begin{Shaded}
\begin{Highlighting}[]
\KeywordTok{from} \NormalTok{pose\_format }\KeywordTok{import} \NormalTok{Pose}
\KeywordTok{from} \NormalTok{pose\_format.pose\_visualizer }\ImportTok{\\}
     \KeywordTok{import} \NormalTok{PoseVisualizer}

\KeywordTok{with} \NormalTok{open(}\StringTok{"example.pose"}\NormalTok{, }\StringTok{"rb"}\NormalTok{) }\KeywordTok{as} \NormalTok{f:}
    \NormalTok{pose = Pose.read(f.read())}

\NormalTok{v = PoseVisualizer(pose)}

\NormalTok{v.save\_video(}\StringTok{"example.mp4"}\NormalTok{, v.draw())}
\end{Highlighting}
\end{Shaded}

\paragraph{Browser}
Additionally, for web-based applications or quick interactive viewing, poses can be visualized in the browser. Unlike the Python visualization, this visualization is vectorized and is more suitable for client-facing applications. 
\begin{Shaded}
\begin{Highlighting}[]
\OperatorTok{<}\KeywordTok{script} \NormalTok{type=}\StringTok{"module"}\OperatorTok{\\}
\NormalTok{src=}\StringTok{"https://unpkg.com/pose-viewer@0.0.1}\OperatorTok{\\}
\NormalTok{/dist/pose-viewer/pose-viewer.esm.js"} \OperatorTok{/>}

\OperatorTok{<}\KeywordTok{pose-format} \NormalTok{src=}\StringTok{"example.pose"} \OperatorTok{/>}
\end{Highlighting}
\end{Shaded}

\section{Benchmarking}
\label{sec:benchmarking}

To evaluate our custom file format, we benchmarked it against OpenPose, a prevalent standard. Metrics of interest were read speed and file size.
We obtained OpenPose data from a single video in the Public DGS Corpus \cite[\footnotesize{DOI: /10.25592/dgs.corpus-3.0-text-1413451-11105600-11163240}]{dataset:hanke-etal-2020-extending}.
Their format employs a monolithic JSON file to store frames, diverging from the common one-file-per-frame approach.

To gauge reading performance, we measured OpenPose's JSON load time in isolation, sidestepping tensor conversion. For our format, we include both full-file reads and body-only tensor reads where we skip loading the pose header, and only load the tensor of coordinates and confidences.

\paragraph{Quantitative Edge}
Table \ref{table:benchmarking} reveals we achieve up to a $60\%$ file size reduction and outpace OpenPose in read speed by a staggering $162\times$, thereby obliterating any machine learning bottlenecks.

\paragraph{Qualitative Edge}
Our \texttt{pose-format} packs all pose data into a singular, robust file, avoiding the file fragmentation issues seen in OpenPose. Moreover, our header encodes pose structure, obviating the need for hard-coded interpretation logic and boosting both portability and usability.

In summation, \texttt{pose-format} offers superior performance across key metrics, making it a compelling alternative for pose data management.

\section{Community Contributions}

Our library is fully open-source, and released under an MIT License. We welcome contributions from the community of any kind, and we encourage collaboration. Source code and bug reporting are available at \url{https://github.com/sign-language-processing/pose}.

\bibliography{anthology,custom}

\begin{thebibliography}{9}
\expandafter\ifx\csname natexlab\endcsname\relax\def\natexlab#1{#1}\fi

\bibitem[{Abadi et~al.(2015)Abadi, Agarwal, Barham, Brevdo, Chen, Citro, Corrado, Davis, Dean, Devin, Ghemawat, Goodfellow, Harp, Irving, Isard, Jia, Jozefowicz, Kaiser, Kudlur, Levenberg, Man\'{e}, Monga, Moore, Murray, Olah, Schuster, Shlens, Steiner, Sutskever, Talwar, Tucker, Vanhoucke, Vasudevan, Vi\'{e}gas, Vinyals, Warden, Wattenberg, Wicke, Yu, and Zheng}]{tensorflow2015-whitepaper}
Mart\'{i}n Abadi, Ashish Agarwal, Paul Barham, Eugene Brevdo, Zhifeng Chen, Craig Citro, Greg~S. Corrado, Andy Davis, Jeffrey Dean, Matthieu Devin, Sanjay Ghemawat, Ian Goodfellow, Andrew Harp, Geoffrey Irving, Michael Isard, Yangqing Jia, Rafal Jozefowicz, Lukasz Kaiser, Manjunath Kudlur, Josh Levenberg, Dandelion Man\'{e}, Rajat Monga, Sherry Moore, Derek Murray, Chris Olah, Mike Schuster, Jonathon Shlens, Benoit Steiner, Ilya Sutskever, Kunal Talwar, Paul Tucker, Vincent Vanhoucke, Vijay Vasudevan, Fernanda Vi\'{e}gas, Oriol Vinyals, Pete Warden, Martin Wattenberg, Martin Wicke, Yuan Yu, and Xiaoqiang Zheng. 2015.
\newblock \href {https://www.tensorflow.org/} {{TensorFlow}: Large-scale machine learning on heterogeneous systems}.
\newblock Software available from tensorflow.org.

\bibitem[{{Cao} et~al.(2019){Cao}, {Hidalgo Martinez}, {Simon}, {Wei}, and {Sheikh}}]{pose:cao2018openpose}
Z.~{Cao}, G.~{Hidalgo Martinez}, T.~{Simon}, S.~{Wei}, and Y.~A. {Sheikh}. 2019.
\newblock {O}pen{P}ose: Realtime multi-person {2D} pose estimation using part affinity fields.
\newblock \emph{IEEE Transactions on Pattern Analysis and Machine Intelligence}.

\bibitem[{Grishchenko and Bazarevsky(2020)}]{MediaPipe2020holistic}
Ivan Grishchenko and Valentin Bazarevsky. 2020.
\newblock \href {https://google.github.io/mediapipe/solutions/holistic.html} {Mediapipe holistic}.

\bibitem[{Hanke et~al.(2020)Hanke, Schulder, Konrad, and Jahn}]{dataset:hanke-etal-2020-extending}
Thomas Hanke, Marc Schulder, Reiner Konrad, and Elena Jahn. 2020.
\newblock \href {https://www.aclweb.org/anthology/2020.signlang-1.12} {Extending the {P}ublic {DGS} {C}orpus in size and depth}.
\newblock In \emph{Proceedings of the LREC2020 9th Workshop on the Representation and Processing of Sign Languages: Sign Language Resources in the Service of the Language Community, Technological Challenges and Application Perspectives}, pages 75--82, Marseille, France. European Language Resources Association (ELRA).

\bibitem[{Harris et~al.(2020)Harris, Millman, van~der Walt, Gommers, Virtanen, Cournapeau, Wieser, Taylor, Berg, Smith, Kern, Picus, Hoyer, van Kerkwijk, Brett, Haldane, del R'io, Wiebe, Peterson, G'erard-Marchant, Sheppard, Reddy, Weckesser, Abbasi, Gohlke, and Oliphant}]{Harris2020ArrayPW}
Charles~R. Harris, K.~Jarrod Millman, St{\'e}fan van~der Walt, Ralf Gommers, Pauli Virtanen, David Cournapeau, Eric Wieser, Julian Taylor, Sebastian Berg, Nathaniel~J. Smith, Robert Kern, Matti Picus, Stephan Hoyer, Marten~Henric van Kerkwijk, Matthew Brett, Allan Haldane, Jaime~Fern'andez del R'io, Marcy Wiebe, Pearu Peterson, Pierre G'erard-Marchant, Kevin Sheppard, Tyler Reddy, Warren Weckesser, Hameer Abbasi, Christoph Gohlke, and Travis~E. Oliphant. 2020.
\newblock \href {https://api.semanticscholar.org/CorpusID:219792763} {Array programming with numpy}.
\newblock \emph{Nature}, 585:357 -- 362.

\bibitem[{Moryossef et~al.(2021)Moryossef, Tsochantaridis, Dinn, Camgoz, Bowden, Jiang, Rios, Muller, and Ebling}]{moryossef2021evaluating}
Amit Moryossef, Ioannis Tsochantaridis, Joe Dinn, Necati~Cihan Camgoz, Richard Bowden, Tao Jiang, Annette Rios, Mathias Muller, and Sarah Ebling. 2021.
\newblock Evaluating the immediate applicability of pose estimation for sign language recognition.
\newblock In \emph{Proceedings of the IEEE/CVF conference on computer vision and pattern recognition}, pages 3434--3440.

\bibitem[{M{\"u}ller et~al.(2022)M{\"u}ller, Ebling, Avramidis, Battisti, Berger, Bowden, Braffort, Cihan~Camg{\"o}z, Espa{\~n}a-bonet, Grundkiewicz, Jiang, Koller, Moryossef, Perrollaz, Reinhard, Rios, Shterionov, Sidler-miserez, and Tissi}]{muller-etal-2022-findings}
Mathias M{\"u}ller, Sarah Ebling, Eleftherios Avramidis, Alessia Battisti, Mich{\`e}le Berger, Richard Bowden, Annelies Braffort, Necati Cihan~Camg{\"o}z, Cristina Espa{\~n}a-bonet, Roman Grundkiewicz, Zifan Jiang, Oscar Koller, Amit Moryossef, Regula Perrollaz, Sabine Reinhard, Annette Rios, Dimitar Shterionov, Sandra Sidler-miserez, and Katja Tissi. 2022.
\newblock \href {https://aclanthology.org/2022.wmt-1.71} {Findings of the first {WMT} shared task on sign language translation ({WMT}-{SLT}22)}.
\newblock In \emph{Proceedings of the Seventh Conference on Machine Translation (WMT)}, pages 744--772, Abu Dhabi, United Arab Emirates (Hybrid). Association for Computational Linguistics.

\bibitem[{Paszke et~al.(2019)Paszke, Gross, Massa, Lerer, Bradbury, Chanan, Killeen, Lin, Gimelshein, Antiga, Desmaison, K{\"o}pf, Yang, DeVito, Raison, Tejani, Chilamkurthy, Steiner, Fang, Bai, and Chintala}]{pytorch}
Adam Paszke, Sam Gross, Francisco Massa, Adam Lerer, James Bradbury, Gregory Chanan, Trevor Killeen, Zeming Lin, Natalia Gimelshein, Luca Antiga, Alban Desmaison, Andreas K{\"o}pf, Edward Yang, Zach DeVito, Martin Raison, Alykhan Tejani, Sasank Chilamkurthy, Benoit Steiner, Lu~Fang, Junjie Bai, and Soumith Chintala. 2019.
\newblock \href {https://api.semanticscholar.org/CorpusID:202786778} {Pytorch: An imperative style, high-performance deep learning library}.
\newblock In \emph{Neural Information Processing Systems}.

\bibitem[{Zheng et~al.(2023)Zheng, Wu, Chen, Yang, Zhu, Shen, Kehtarnavaz, and Shah}]{zheng2023deep}
Ce~Zheng, Wenhan Wu, Chen Chen, Taojiannan Yang, Sijie Zhu, Ju~Shen, Nasser Kehtarnavaz, and Mubarak Shah. 2023.
\newblock \href {https://doi.org/10.1145/3603618} {Deep learning-based human pose estimation: A survey}.
\newblock \emph{ACM Comput. Surv.}, 56(1).

\end{thebibliography}
\bibliographystyle{acl_natbib}

\end{document}